\documentclass[11pt,a4paper]{article}
\usepackage[hyperref]{acl2020}
\usepackage{times}
\usepackage{latexsym}

\usepackage{microtype}

\usepackage{graphicx}
\usepackage{amsmath}
\usepackage{amssymb}
\usepackage{bm}
\usepackage{multirow}
\usepackage{comment}

\DeclareMathOperator*{\argmax}{arg\,max}
\DeclareMathOperator*{\argmin}{arg\,min}

\aclfinalcopy 

\title{Improved Synthetic Training for Reading Comprehension}

\author{
Yanda Chen$^{\dag}$\thanks{\hspace{2mm}Work done during summer internship at IBM Research.}~~~
Md Arafat Sultan$^{\ddag}$~~~
Vittorio Castelli$^{\ddag}$ \\
$^{\dag}$Department of Computer Science, Columbia University \\
$^{\ddag}$IBM Research AI, T.J. Watson Research Center, New York, USA \\
{\fontsize{11}{11} \selectfont {\tt {yc3384@columbia.edu,arafat.sultan@ibm.com,vittorio@us.ibm.com}}}
}

\date{}

\begin{document}
\maketitle
\begin{abstract}
Automatically generated \textit{synthetic} training examples have been shown to improve performance in machine reading comprehension (MRC).
Compared to human annotated gold standard data, synthetic training data has unique properties, such as high availability at the possible expense of quality. 
In view of such differences, in this paper, we explore novel applications of synthetic examples to MRC.
Our proposed pre-training and knowledge distillation strategies show significant improvements over existing methods.
In a particularly surprising discovery, we observe that synthetic distillation often yields students that can outperform the teacher model. 
\end{abstract}

\section{Introduction}
Recent advances in text-to-text generation have enabled significant progress in automatic creation of training examples for machine reading comprehension (MRC) \cite{dong2019unified,alberti-etal-2019-synthetic,sultan-etal-2020-importance}.
The advent of massive pre-trained language models \cite{Radford:2019,lewis-etal-2020-bart} alongside improved decoding techniques for
open-ended text generation \cite{Holtzman:2019} has largely reduced the generation of synthetic examples to two simple steps: (1) fine-tuning of a language model on an existing MRC dataset \cite{dong2019unified,alberti-etal-2019-synthetic}, and (2) employing an appropriate decoder for generation \cite{sultan-etal-2020-importance}.

However, less is understood about how synthetic MRC examples can be best utilized, possibly in ways that go beyond standard direct supervision.
Machine generated examples are different than gold standard human annotations along two key dimensions: (a)~they are noisier, and (b)~they can be generated in numbers that are orders of magnitude larger.
In this paper, we explore two novel applications of synthetic MRC examples that exploit their abundance while relying on improved denoising techniques.

\begin{figure}[t]
\centering
\includegraphics[width=7.8cm]{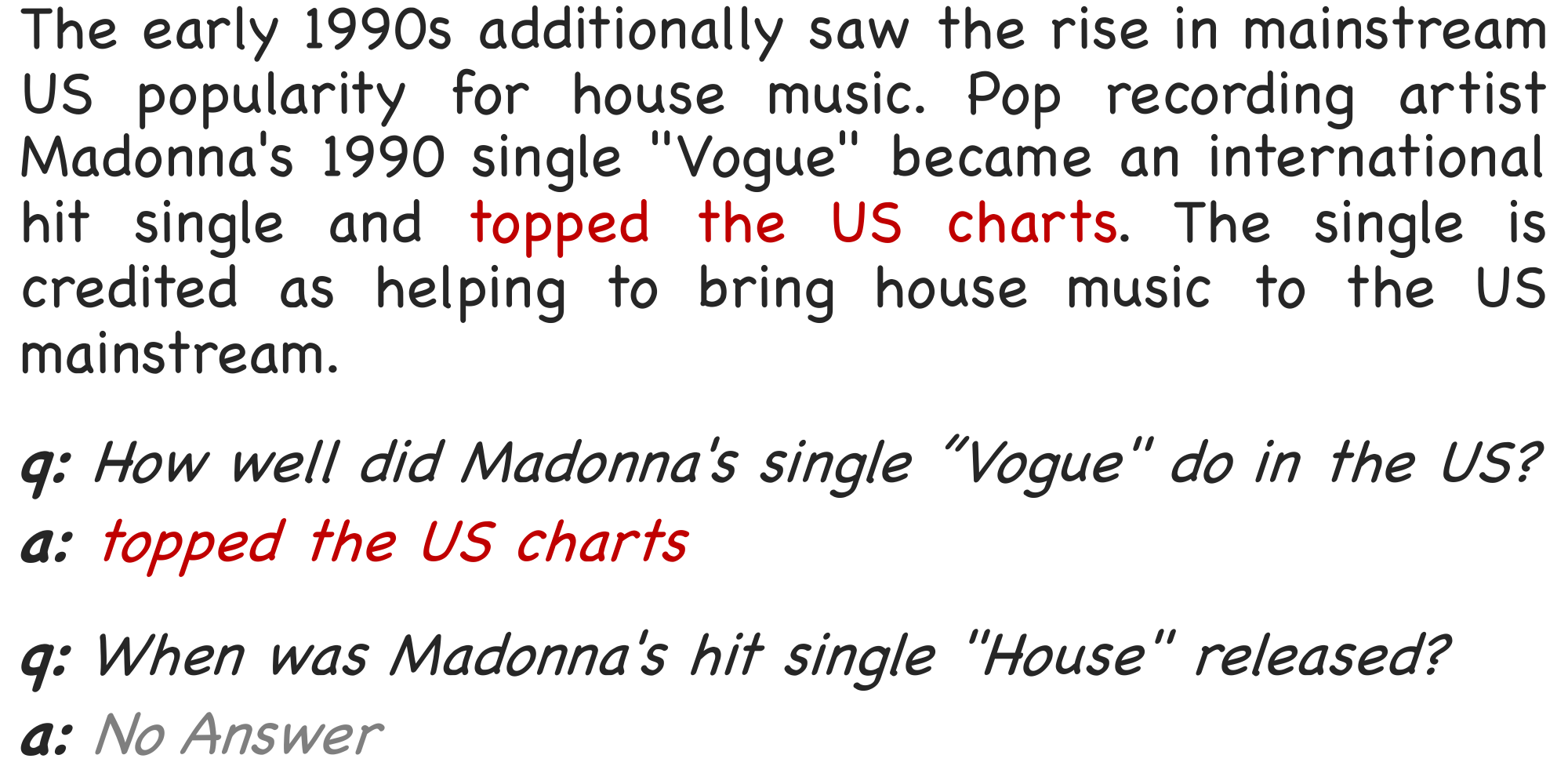}
\caption{MRC examples from SQuAD2.0 \cite{rajpurkar-etal-2018-know}. The first question has an answer in the context (top) but the second question does not.}
\label{figure:mrc-examples}
\end{figure}

We study a common form of extractive MRC (see Figure~\ref{figure:mrc-examples}), where given a textual context $c$ and a question $q$, the system must first determine if $q$ has an answer in $c$, and then extract the answer $a$ if $q$ is answerable.
Given $c$, generation of a synthetic example thus requires both asking a question $q$ and answering it with either a span $a$ in $c$ or \textit{``No Answer''}.

A commonly used strategy to deal with noise in artificial training data is synthetic pre-training: the MRC model is pre-trained on synthetic examples before being fine-tuned on human annotated examples \cite{dong2019unified,sultan-etal-2020-importance}.
\citet{alberti-etal-2019-synthetic} also show that verification of ``roundtrip consistency'' of synthetic examples using a trained MRC model can improve performance over standard pre-training.
Here we propose the use of synthetic pre-training to specifically target known weaknesses of an existing MRC model.
We show that roundtrip-consistent examples for which a given MRC model has high prediction loss can be highly effective pre-training examples for that model.
In our experiments with two benchmark MRC datasets, namely SQuAD2.0 \cite{rajpurkar-etal-2018-know} and NewsQA \cite{Trischler:2017}, smaller subsets of such high-loss examples yield better MRC training than larger populations of random roundtrip-consistent examples.

The second proposed application, where our findings are arguably more far-reaching, involves knowledge distillation \cite{hinton2015distilling}.
It is well established in the literature that distillation from a high-performance teacher model can help improve performance of a weaker student model on target tasks \cite{liu2019improving,sanh2020distilbert}.
We hypothesize that synthetic examples generated in large numbers can uncover the teacher's knowledge in greater detail, enabling better distillation than limited amounts of gold standard data.
Our experiments on the above benchmarks with a \textsc{BERT-Large} teacher (340\textsc{m} parameters) and a \textsc{BERT-Base} student (110\textsc{m} parameters) confirm this hypothesis.
To our surprise, further distillation with human annotated examples often elevates the student's performance above that of the teacher, even when the teacher is synthetically pre-trained for improved performance.

These findings have major implications for real-life application of QA systems, as smaller models are often desired for their low cost and high speed.
Moreover, while we focus on MRC in this paper, our proposed ideas are generally applicable to all supervised tasks for which synthetic examples can be generated.

\section{Related Work}
Starting from early rule-based approaches that relied on syntactic transformations or handcrafted semantic templates \cite{heilman-smith-2010-good, lindberg-etal-2013-generating, mazidi-nielsen-2014-linguistic}, automatic question generation from text has gradually transitioned to neural sequence-to-sequence generation methods \cite{du-etal-2017-learning,duan-etal-2017-question, harrison-walker-2018-neural, zhu-etal-2019-learning}.
Most state-of-the-art generators also benefit from large-scale language model pre-training \cite{dong2019unified, scialom-etal-2019-self, Liu_2020}.

A number of recent studies have demonstrated the utility of synthetic examples in MRC model training \cite{duan-etal-2017-question, sachan-xing-2018-self, zhang-bansal-2019-addressing}. 
While prior work has largely focused on generating high-quality examples measured by properties such as accuracy \cite{alberti-etal-2019-synthetic, Liu_2020, dong2019unified} and diversity \cite{sultan-etal-2020-importance}, our main goal in this paper is effective post-generation selection and application of synthetic examples.


Identification of informative examples is a critical subproblem in active learning, where an oracle is queried with such examples for labels.
Various query strategies have been proposed, which include uncertainty sampling \cite{10.5555/188490.188495, 10.5555/647967.741626, Wang_2017, gal2017deep}, query-by-committee \cite{10.1145/130385.130417, ghai2020active, chae2020poolbased} and expected error reduction \cite{10.5555/645530.655646, konyushkova2017learning}.
Among these, uncertainty sampling methods select examples for which the model being trained is the least certain about what the output should be.
We share the general goal of identifying the most useful training examples, but instead of querying an oracle based on model uncertainty, we sample from an existing pool of synthetic examples based on model error.

Another related approach is core-set selection, which attempts to find a representative subset of examples that accurately approximates a larger dataset \cite{10.1145/1064092.1064114, huggins2017coresets, sener2018active, coleman2020selection}. 
While our goal is also to identify a useful subset of examples, rather than approximating the entire synthetic dataset, we intend to identify pre-training examples that can augment human annotated examples well.

Finally, knowledge distillation \cite{hinton2015distilling} has proven to be an extremely useful technique for building small yet high-performance models by training them to imitate larger teacher models.
Various generalizations have been proposed, such as multi-task distillation \cite{liu2019improving}, multi-teacher distillation \cite{yang2019model} and distillation with hidden layer supervision \cite{romero2015fitnets, sun-etal-2019-patient}.


Previous research has shown that language model pre-training on large amounts of unlabeled text, via language model distillation \cite{sanh2020distilbert} or standard language model pre-training followed by task-specific distillation \cite{turc2019wellread}, can improve student performance.
Our work is perhaps more closely related to the two-stage distillation framework presented in \cite{yang2019model}, where large amounts of question-passage pairs are derived from a commercial web search engine to create a binary relevance judgment pre-training task.
Student models are distilled first on this pre-training task and later on target task examples.
Unlike all these approaches, we rely on automatically generated target task (MRC) examples for effective distillation.


\section{Synthetic Training Methods}
In this section, we discuss the generation of synthetic MRC examples as well as our proposed strategies for their application in pre-training and knowledge distillation.

\subsection{Example Generation}
\label{sec:example-generation}

Following \textsc{anonymized}, we fine-tune an autoregressive encoder-decoder language model \cite{lewis-etal-2020-bart} to generate synthetic MRC examples with answerable questions.
Let $c$ be a paragraph in a document $d$, $q$ a question, and $a$ its answer in $c$.
Let $s$ be the sentence in $c$ that contains $a$.
Our generator is trained to model the conditional joint distribution $p(s,a,q|c)$, which in essence enables it to simultaneously identify an appropriate candidate $s$ in $c$, extract $a$ from $s$, and generate $q$.
In practice, in the generated sequence $(s,a,q)$, we only include the first and the last word of $s$ to represent $s$.
Special separator tokens are used to separate the three elements of the generated triple.

Given a dataset $D$ of answerable MRC examples, the parameters $\theta$ of the generator are learned using standard maximum likelihood estimation:

\begin{equation*}
    \theta^{*} = \argmax_{\theta} \sum_{i=1}^{|D|} \log p_{\theta}(s_i, a_i, q_i \mid c_i)
\end{equation*}

At inference time, given a document $d$, we first generate a triple $(s, a, q)$ from a paragraph $c$ in $d$ using top-$p$ top-$k$ sampling. 
We retain $q$ and $a$ to create an answerable MRC example $(c, q, a)$.
Finally, to create an unanswerable example for $q$, we follow \citet{alberti-etal-2019-synthetic} to simply pair up $q$ with a different paragraph $c'$ in $d$ which results in the example $(c', q, \textit{``No Answer''})$.


\subsection{MRC Training}
\label{sec:mrc-training}
Following \citet{Devlin:2019}, we fine-tune a transformer-based masked language model (LM)
for MRC.
This section provides an overview of the procedure; we refer the reader to \cite{Devlin:2019} for further details.

Let $c$ be a context, $q$ a question, and $a$ its answer in $c$.
Let $a_{start}$ and $a_{end}$ be the start and end offsets of $a$ in $c$.
The input to the MRC system is a concatenation of $q$ and $c$, separated by a special separator token.
The MRC network consists of two fully connected feed-forward subnetworks atop shared LM transformer layers, which learn to predict the start and end probabilities $p_s(a_{start}|q, c)$ and $p_e({a_{end}|q, c})$, respectively.

Given a dataset $D$ of MRC examples, the parameters $\theta$ of the MRC system are learned using standard maximum likelihood estimation:  
\begin{align*}
    \theta^{*} &= \argmax_{\theta} \sum_{i=1}^{|D|} \log p_{\theta}(a_i \mid q_i, c_i)\\
    &= \argmax_{\theta} \sum_{i=1}^{|D|} \big\{\log p_{s,\theta}(a_{start,i}| q_i, c_i) \\
    &~~~~~~~~~~~~~~~~~~~~~~~~~~~+ \log p_{e,\theta}(a_{end,i}| q_i, c_i)\big\}
\end{align*}

At inference time, the model outputs the answer span $[j:k]$ such that:
\begin{align*}
    j &= \argmax_{j'} p_{s,\theta}(j' \mid q,c) \\
    k &= \argmax_{k'} p_{e,\theta}(k' \mid q,c)
\end{align*}

\subsection{Synthetic Pre-training}
After generating synthetic examples using the generator of Section~\ref{sec:example-generation}, we perform a denoising roundtrip consistency check \cite{alberti-etal-2019-synthetic} using an MRC model separately trained on human annotated gold standard data.
This step removes examples for which the MRC model predicts a different answer than the one in the example.
Let $S$ be the set of roundtrip consistent synthetic examples and $G$ a given set of gold examples.

Following prior work \cite{dong2019unified,sultan-etal-2020-importance}, we adopt a two-step process for the application of $S$ in conjunction with $G$: (1) pre-train the MRC model on $S$, and (2) fine-tune on $G$.
In the rest of this section, we will denote this model by $\theta_{S \rightarrow G}$ and the same network fine-tuned only on gold data by $\theta_G$.

\subsection{Targeted Synthetic Pre-training}
\label{sec:targeted-pretraining}
Given cycle-consistent synthetic training examples $S$, standard pre-training utilizes the entire set $S$ to maximize the amount of training data.
In this section, we propose an approach to identify a subset \mbox{$S' \subset S$} which explicitly encodes the weaknesses of $\theta_G$, and thereby facilitates targeted supervision of a model $\theta_{S' \rightarrow G}$ that is superior to $\theta_{S \rightarrow G}$.


Drawing inspiration from uncertainty sampling techniques in active learning \cite{10.5555/188490.188495, gal2017deep}, we propose \textit{Highest Error} synthetic pre-training, and show that synthetic training examples for which the prediction errors of model $\theta_G$ are the highest provide the best pre-training for $\theta_G$.
More concretely, we define an example difficulty function $H: S \rightarrow \mathbb{R}_{\geq 0}$ for training example $s = (c, q, a) \in S$ as the negative log-likelihood loss of $\theta_G$ for $s$:
\begin{equation*}
    H(s) = -\log p_{\theta_G}(a \mid q, c)
\end{equation*}
Let $S^*$ be the examples in $S$ sorted from hardest to easiest based on their $H$ values :
\begin{equation*}
    S^* = sort(S, \mbox{key=$H$}, \mbox{order=decreasing})
\end{equation*}
To investigate the relationship between example difficulty and pre-training effectiveness, 
we partition $S^*$ into consecutive bins of uniform size $b$. 
Let $n$ be the number of bins so that $b = |S^*|/n$. 
For $i \in \{1, 2, ..., n\}$, the $i$-th bin $B_i$ consists of examples $S^*[(i-1)\times b:i \times b]$ in a randomized order. 
For each bin $B_i \subset S^*$, we train an MRC model $\theta_{B_i \rightarrow G}$ and evaluate the pre-training effectiveness of subset $B_i$ based on the performance of $\theta_{B_i \rightarrow G}$ on a test set. 

\subsection{Synthetic Distillation}
\label{sec:syn-distillation}
Automatically generated synthetic MRC examples often contain noisy labels, i.e., $(c, q, a)$ triples can be generated where $a$ is not a correct answer to $q$ given $c$. 
Such label noise can hurt model performance if maximum likelihood estimation is used for training (Section~\ref{sec:mrc-training}). 
Knowledge distillation \cite{hinton2015distilling}, on the 
other hand, ignores labels in training examples altogether and instead obtains soft labels (probability distributions over possible answers) from a stronger teacher model.
In addition to providing powerful supervision, distillation can thus be a useful denoising operator for synthetic examples. 


We further posit that distillation can mutually benefit from the use of synthetic training data.
Synthetic examples can be generated in large numbers and also with high diversity when an appropriate decoding technique such as top-$p$ top-$k$ sampling is used \cite{sultan-etal-2020-importance}.
Distillation aims to uncover the behavior of the teacher model across a range of input scenarios; we hypothesize that large amounts of diverse synthetic examples can achieve this objective more effectively than limited amounts of human annotated examples.

To test this hypothesis, we perform distillation as follows.
For a training example $(c, q, a)$ in synthetic dataset $S$, let $L$ be the length of the concatenated MRC input $(q, c)$ (see Section~\ref{sec:mrc-training} for details).
Let $z^t_{start}$ and $z^t_{end}$ be the probability distributions of answer start and end offsets, respectively, as predicted by the teacher model over all $L$ positions of the input sequence.
Similarly, let $z_{start}$ and $z_{end}$ be the distributions predicted by the student.
We compute the following distillation loss based on the Kullback-Leibler divergence from $z$ to $z^t$:
\begin{align*}
    \mathcal{L}_{distill, start} &:= \sum_{i=1}^{|S|}D_{KL}(z^t_{start, i}~\|~z_{start, i})\\
    \mathcal{L}_{distill, end} &:= \sum_{i=1}^{|S|}D_{KL}(z^t_{end, i}~\|~z_{end, i})\\
    \mathcal{L}_{distill} &:= \frac{1}{2}(\mathcal{L}_{distill, start} + \mathcal{L}_{distill, end})
\end{align*}
We train the student model by minimizing $\mathcal{L}_{distill}$:
\begin{equation*}
    \theta^{*} = \argmin_{\theta} \mathcal{L}_{distill}
\end{equation*}
At inference time, prediction follows the same procedure as in Section~\ref{sec:mrc-training}.

\section{Experimental Setup}
\label{sec:setup}
In this section we describe our general experimental setup.
Further details specific to individual experiments are provided in Section~\ref{sec:results}.

\subsection{Datasets}
We use two public MRC benchmark datasets in our experiments: SQuAD2.0 \cite{rajpurkar-etal-2018-know} and NewsQA \cite{Trischler:2017}.
The documents in SQuAD2.0 are Wikipedia articles, while NewsQA consists of CNN news arcles.
Since the official Test set of SQuAD2.0 is not publicly available, we use the official Dev set as our Test set, and a random split of the original training documents as Train and Dev.
For NewsQA, we use the official Train-Dev-Test split. Statistics for both datasets are provided in Table~\ref{table:dataset-stats}.

\begin{table}[h]
\fontsize{9.5}{12}\selectfont
\centering
\begin{tabular}{lccc}
\hline
 & \textbf{Train} & \textbf{Dev} & \textbf{Test} \\
\hline
\textbf{SQuAD2.0}  \\
~~$\#$ of Documents & 397 & 45 &  35\\
~~$\#$ of Paragraphs & 17,081 & 1,954 & 1,204 \\
~~$\#$ of Examples & 117,159 & 13,160 & 11,873 \\
\hline
\textbf{NewsQA} \\
~$\#$ of Documents & 11,469 & 638 & 637 \\
~$\#$ of Examples & 107,669 & 5,988 & 5,971  \\
\hline
\end{tabular}
\caption{\label{table:dataset-stats} Dataset statistics. Examples are aligned to paragraphs in SQuAD2.0, but not in NewsQA. 
}
\end{table}

\subsection{Models}
\label{subsec:models}
We fine-tune a \textsc{BART-Large} \cite{lewis-etal-2020-bart} encoder-decoder language model for MRC example generation.
We fine-tune \textsc{BERT-Large} masked language models (340\textsc{m} parameters) \cite{Devlin:2019} for all our synthetic pre-training experiments.
In knowledge distillation experiments, we use \textsc{BERT-Large} teachers and \textsc{BERT-Base} (110\textsc{m} parameters) students.
All our implementations are based on the Hugging Face library of transformers \cite{Wolf:2019}.

\subsection{Synthetic Example Generation}
We train separate generators for SQuAD2.0 and NewsQA on answerable training examples.
At inference time, we use top-$p$ top-$k$ sampling with $p=0.9$ and $k=10$.
Given a generated answerable example $(c, q, a)$, an unanswerable example is created by pairing up $q$ with a randomly sampled context $c' \neq c$ from the same document.

For both datasets, we generate synthetic examples (Section~\ref{sec:example-generation}) from unlabeled in-domain documents.
For SQuAD2.0, these are Wikipedia documents taken from the Natural Questions (NQ) dataset \cite{kwiatkowski-etal-2019-natural}.
For NewsQA, we use two different sources: (1) CNN articles \cite{hermann2015teaching} that are not in NewsQA, and (2) New York Times (NYT) articles in the Gigaword corpus \cite{graff2005english}.
Statistics are shown in Table~\ref{table:syn-example-sources}.

\begin{table}[h]
\fontsize{9.5}{12}\selectfont
\centering
\begin{tabular}{lcc}
\hline
 & \textbf{Wikipedia} & \textbf{CNN + NYT} \\
\hline
$\#$ of Documents & 307,373 & 1,333,316 \\
$\#$ of Paragraphs & 1,812,843 & 4,841,721 \\
\hline
\end{tabular}
\caption{\label{table:syn-example-sources} Statistics of unlabeled corpora from which we generate synthetic examples. Wikipedia articles are used to generate examples for SQuAD2.0; CNN and New York Times (NYT) articles are used for NewsQA.
}
\end{table}

Many CNN and NYT paragraphs are relatively short; we merge these paragraphs to create contexts that are around 320 word pieces long.
Longer paragraphs are used as is.
For SQuAD2.0, individual paragraphs are used as contexts.
We generate five examples per context for SQuAD2.0 and three per context for NewsQA, and remove all duplicates.
We also create one-fourth as many unanswerable examples as answerable ones for each dataset. 

For roundtrip consistency check, we propose to utilize a different, preferably stronger model than the one we pre-train with synthetic examples.
While other options such as an ensemble of different models exist, we simply train \textsc{RoBERTa-Large} MRC models on the respective gold datasets as consistency checkers of synthetic examples.
Following \citet{alberti-etal-2019-synthetic}, we only use roundtrip consistent examples for synthetic pre-training.
For each dataset, we finally retain a random sample of \textsc{4m} answerable and \textsc{1m} unanswerable roundtrip consistent examples to use in our experiments. Table~\ref{table:syn-examples-stats} shows the statistics. 

\begin{table}[t]
\fontsize{9.2}{12}\selectfont
\centering
\begin{tabular}{lcc}
\hline
 & \textbf{SQuAD2.0} & \textbf{NewsQA} \\
\hline
Total Answerable & 7,574,159 & 12,389,865 \\
Total Unanswerable & 1,900,000 & 3,097,466 \\
RC Answerable & 4,954,869 & 4,753,991 \\
RC Unanswerable & 1,788,476 & 2,978,017 \\
RC Answerable in \textsc{Syn} & \textsc{4m} & \textsc{4m} \\
RC Unanswerable in \textsc{Syn} & \textsc{1m} & \textsc{1m} \\
\hline
\end{tabular}
\caption{\label{table:syn-examples-stats} Counts of automatically generated examples.
RC: Roundtrip Consistent; \textsc{Syn}: final set of pre-training examples used in our experiments. 
}
\end{table}

\begin{table}[h]
\fontsize{8.2}{11}\selectfont
\centering
\begin{tabular}{lcccc}
 & & \textbf{SQuAD2.0} & \textbf{NewsQA} \\
\hline
Batch Size & & 12 & 24 \\
\hline
\multirow{2}{*}{LR} & \textsc{BERT-Base} & 3$\times 10^{-5}$ & 3$\times 10^{-5}$ \\
& \textsc{BERT-Large} & 3$\times 10^{-5}$ & 2$\times 10^{-5}$ \\
\hline
\multirow{4}{*}{$\#$ of Epochs} & \textsc{Gld Hl} & 2 & 1 \\
& \textsc{Syn $\rightarrow$ Gld Hl} & 1 $\rightarrow$ 2 & 1$\rightarrow$ 1 \\
& \textsc{Gld Dst} & 6 & 4 \\
& \textsc{Syn $\rightarrow$ Gld Dst} & 1 $\rightarrow$ 4 & 1 $\rightarrow$ 4 \\
\hline
\end{tabular}
\caption{\label{table:training-config} MRC training configurations. LR: learning rate; \textsc{Hl}: hard label training (Section~\ref{sec:mrc-training}); \textsc{Dst}: soft distillation training (Section~\ref{sec:syn-distillation}). 
}
\end{table}

\subsection{MRC Training}
Training configurations of the MRC models for SQuAD2.0 and NewsQA are shown in Table~\ref{table:training-config}.
The SQuAD2.0 configuration is the Hugging Face default.
For NewsQA, we observe that performance is more sensitive to the training configuration.
Hence we conduct a hyperparameter grid search for NewsQA with a model trained only on gold examples and choose the set of values that yield the best Dev performance.  

\subsection{Knowledge Distillation}
In order to train a high performance \mbox{\textsc{BERT-Base}} student, for each dataset, we select as our teacher model the respective best performing \mbox{\textsc{BERT-Large}} model from Section~\ref{sec:targeted-pretraining}, trained using highest error synthetic pre-training.

\subsection{Evaluation}
We report results on both Dev and Test sets of the two benchmarks.
Unless otherwise specified, the reported numbers are overall F1 scores that include both answerable and unanswerable examples. 
We report best $F_1$ scores on Dev, and use the best Dev threshold to compute $F_1$ scores on Test.
We refer the reader to \cite{rajpurkar-etal-2018-know} for details on these metrics.

\section{Results and Analysis}
\label{sec:results}
We run all our experiments with three different random seeds.
All numbers reported in this section are averages over the three seeds.

\subsection{Targeted Synthetic Pre-training}
\label{sec:synth-pretrain-results}
\begin{figure}
\centering
\includegraphics[width=8.5cm,trim=1.5cm 0 0 1.5cm]{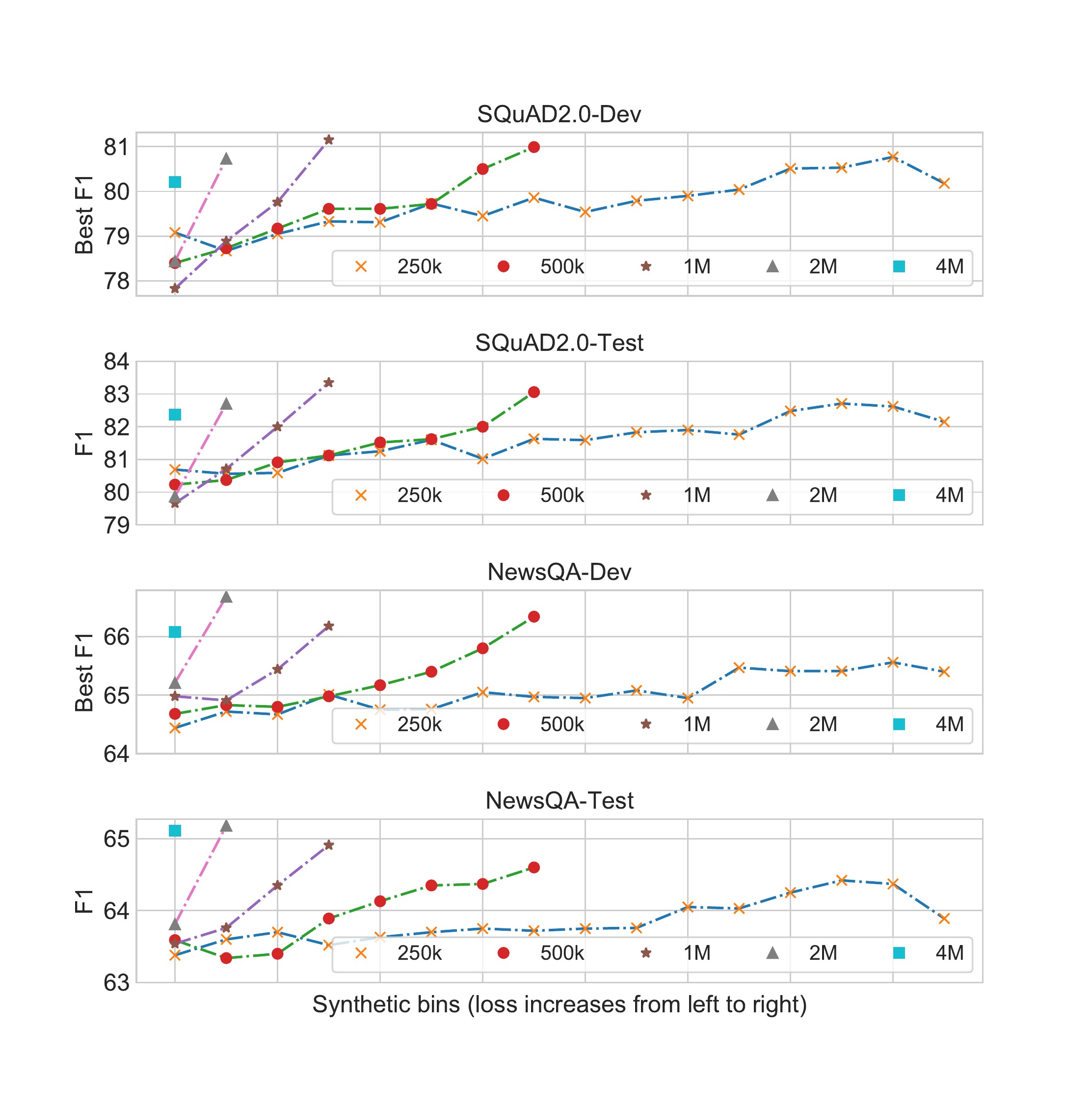}
\caption{Pre-training results for different difficulty bins of synthetic examples. Each curve represents a bin size ranging from 250k to 2\textsc{m} answerable examples. Each bin also contains one-fourth as many unanswerable examples. The square represents the entire population of 4\textsc{m} answerable and \textsc{1m} unanswerable examples.}
\label{figure:synbins}
\end{figure}

\begin{table}[t]
\fontsize{9}{11}\selectfont
\centering
\begin{tabular}{lcccc}
\multicolumn{1}{c}{} & \multicolumn{2}{c}{\textbf{SQuAD2.0}} & \multicolumn{2}{c}{\textbf{NewsQA}} \\
\cline{2-5}
Bin Size & Dev & Test & Dev & Test \\
 \hline
250k & 1.1 & 1.5 & 1.0 & 0.5 \\
500k & 2.6 & 2.8 & 1.7 & 1.0 \\
\textsc{1m} & 3.3 & 3.7 & 1.2 & 1.4 \\
\textsc{2m} & 2.3 & 2.8 & 1.5 & 1.4 \\
\hline 
\end{tabular}
\caption{\label{table:pretraining-difference-results} Performance difference ($F_1$ scores) between the models pre-trained on the hardest and the easiest bins for different bin sizes. 
}
\end{table}

To examine the relationship between difficulty and pre-training effectiveness of synthetic examples, we follow the procedure in Section~\ref{sec:targeted-pretraining} to first partition the \textsc{4m} answerable and \textsc{1m} unanswerable synthetic examples into bins of different difficulty levels.
We experiment with different bin sizes: 250k (16 bins), 500k (8 bins), \textsc{1m} (4 bins), \textsc{2m} (2 bins), and all \textsc{4m} (1 bin) answerable examples, plus one-fourth as many unanswerable examples per bin. 
For each bin $B \subset S$, we fine-tune a \textsc{BERT-Large} MRC model $\theta_{B \rightarrow G}$ as discussed in Section~\ref{sec:mrc-training}. Performances of models trained on different bins are shown in Figure~\ref{figure:synbins}.

\begin{table}[b]
\fontsize{8.1}{11}\selectfont
\centering
\begin{tabular}{lcccc}
\multicolumn{1}{c}{} & \multicolumn{2}{c}{\textbf{SQuAD2.0}} & \multicolumn{2}{c}{\textbf{NewsQA}} \\
\cline{2-5}
 & Dev & Test & Dev & Test \\
 \hline
\textsc{Gld} & 79.2 & 80.9 & 64.3 & 63.2 \\
\textsc{Syn 4m $\rightarrow$ Gld} & 80.2 & 82.4 & 66.1 & 65.1 \\
\hline 
\textsc{Syn} 250k Hard $\rightarrow$ \textsc{Gld} & 80.2 & 82.2 & 65.4 & 63.9 \\
\textsc{Syn} 500k Hard $\rightarrow$ \textsc{Gld} & ~81.0\textsuperscript{$\dagger$} & ~83.1\textsuperscript{$\dagger$} & ~66.3\textsuperscript{$\dagger$} & 64.6  \\
\textsc{Syn 1m} Hard $\rightarrow$ \textsc{Gld} & ~\textbf{81.1}\textsuperscript{$\dagger$} & ~\textbf{83.3}\textsuperscript{$\dagger$} &  ~66.2\textsuperscript{$\dagger$} & 64.9 \\
\textsc{Syn 2m} Hard $\rightarrow$ \textsc{Gld} & ~80.7\textsuperscript{$\dagger$} & ~82.7\textsuperscript{$\dagger$} & ~\textbf{66.7}\textsuperscript{$\dagger$} & ~\textbf{65.2}\textsuperscript{$\dagger$} \\
\hline
\end{tabular}
\caption{\label{table:pretraining-main-results} Performance ($F_1$ scores) of \textsc{BERT-Large} with gold-only training (\textsc{Gld}) and different synthetic pre-training (\textsc{Syn}) subsets. \textsuperscript{$\dagger$} indicates subsets that are better than the entire \textsc{4m} collection. 
}
\end{table}

For bin sizes between 500k and \textsc{2m} answerable examples (inclusive), we observe that model performance consistently improves from easier bins (examples with smaller losses) to harder bins (examples with larger losses) on both Dev and Test of SQuAD2.0 and NewsQA.
We also observe that this pattern, while still present, is weaker and noisier for the 250k bin.
Let \textsc{Hard} and \textsc{Easy} denote the hardest and the easiest bin for each bin size. 
In Table~\ref{table:pretraining-difference-results}, we report the performance difference between $\theta_{\textsc{Hard} \rightarrow G}$ and $\theta_{\textsc{Easy} \rightarrow G}$: the former consistently outperforms the latter across all bin sizes and test conditions.
The above results confirm that harder examples are generally better pre-training examples than easier examples.

Next we examine the relative performance of the set of all synthetic examples $S$ (i.e., the largest \textsc{4m} bin) against the smaller bins.
Table~\ref{table:pretraining-main-results} shows that in all four test conditions (SQuAD2.0 Dev/Test, NewsQA Dev/Test), higher performance can be achieved by pre-training only on a hard subset of $S$ rather than on all of $S$. In fact, we see that in all test conditions except NewsQA Test, the hardest 500k bin already provides better pre-training than $S$, reducing the amount of synthetic examples needed by 87.5\%. However, in none of the four test conditions is the hardest 250k bin as good as $S$, implying that even though hardest examples are the most effective, enough of them must be included in pre-training to ensure sufficient sample diversity. 

\begin{table}[t]
\fontsize{9}{11}\selectfont
\centering
\begin{tabular}{lcccc}
\multicolumn{1}{c}{} & \multicolumn{2}{c}{\textbf{SQuAD2.0}} & \multicolumn{2}{c}{\textbf{NewsQA}} \\
\cline{2-5}
 & Dev & Test & Dev & Test \\
 \hline
Plain CL & 78.2 & 80.0 & 65.7 & 64.6 \\
CL with 5\% Switch & 78.8 & 80.8 & 65.8 & 64.4 \\
Random Training Order & \textbf{80.2} & \textbf{82.4} & \textbf{66.1} & \textbf{65.1} \\
\hline
\end{tabular}
\caption{\label{table:cl-results} Performance of curriculum learning is worse than a randomized training order.
}
\end{table}

Finally, the proposed idea of using the hardest subset of examples for pre-training has similarities with curriculum learning (CL) \cite{Bengio:2009}, which trains models in an easy-to-hard order so that the hardest examples are used at the end.
In Table~\ref{table:cl-results}, we show results for pre-training on all examples $S$ with CL.
As prior work suggests that introducing some harder examples early in CL can be useful \cite{platanios2019competencebased, penha2019curriculum}, we also examine a version of CL where positions of 5\% of the examples are randomly switched. 
As Table~\ref{table:cl-results} shows, neither version of CL performs as well as a completely randomized order, which is the baseline model in Table~\ref{table:pretraining-main-results} (row 2).
Note that in our proposed method, examples in each bin are internally randomly ordered (Section~\ref{sec:targeted-pretraining}).
We conjecture that in CL, training on easier examples first might confine the model's parameters to a region that later on makes generalization difficult.

\subsection{Synthetic Distillation}
In our knowledge distillation experiments, we use the best \textsc{BERT-Large} model for each dataset from Section~\ref{sec:synth-pretrain-results} as the teacher model: \textsc{Syn 1m} Hard $\rightarrow$ \textsc{Gld} for SQuAD2.0 and \textsc{Syn 2m} Hard $\rightarrow$ \textsc{Gld} for NewsQA (see Table~\ref{table:pretraining-main-results}).
We train \textsc{BERT-Base} students for all experiments.

Given a training dataset $D$, knowledge distillation can generally use a combination of a distillation loss such as $\mathcal{L}_{distill}$ from Section~\ref{sec:syn-distillation} and a standard negative log-likelihood loss based on hard labels from $D$:

\begin{equation*}
    \mathcal{L}_{hard} = - \sum_{i=1}^{|D|} \log p_\theta (a_i \mid q_i, c_i)
\end{equation*}
To find out the best combination for our models, we first train several students on the gold training sets, for different values of $\lambda \in [0, 1]$ in the following joint loss:
\begin{equation*}
    \mathcal{L} = \lambda \mathcal{L}_{distill} + (1 - \lambda)\mathcal{L}_{hard} 
\end{equation*}
SQuAD2.0 models are trained for 6 epochs and NewsQA models for 4 epochs (tuned on Dev).

Results are shown in Table~\ref{table:gold-disitillation-results}.
The best gold distillation 
results are achieved with $\lambda=1.0$, where training discards hard labels entirely and only uses logits predicted by the teacher model.
On the other hand, hard label MLE training ($\lambda=0.0$) has the lowest $F_1$ scores, indicating that any amount of distillation is useful over using only hard labels.
We use only $\mathcal{L}_{distill}$ in all later experiments.

Expectedly, but importantly, there is a performance gap between the teacher and the best student model for all test conditions in Table~\ref{table:gold-disitillation-results} (more in SQuAD2.0 than in NewsQA).
We posit that this gap is likely a function of the limited availability of human annotated gold examples, which do not provide enough sample diversity to fully uncover the behavior of the teacher model in a range of input scenarios.



\begin{table}[t]
\fontsize{8.6}{10}\selectfont
\centering
\begin{tabular}{lcccc}
\multicolumn{1}{c}{} & \multicolumn{2}{c}{\textbf{SQuAD2.0}} & \multicolumn{2}{c}{\textbf{NewsQA}} \\
\cline{2-5}
& Dev & Test & Dev & Test \\
 \hline
$\lambda=0.0$ \textsc{(Gold MLE)} & 74.7 & 75.8 & 62.2 & 60.6 \\
$\lambda=0.3$ & 76.6 & 77.7 & 63.7 & 63.0 \\
$\lambda=0.5$ & 77.0 & 77.9 & 64.6 & 64.0 \\
$\lambda=0.7$ & 77.2 & 78.3 & 65.6 & 64.6 \\
$\lambda=0.9$ & \textbf{77.4} & 78.5 & \textbf{66.0} & 64.7 \\
$\lambda=1.0$ (\textsc{Soft Labels}) & \textbf{77.4} & \textbf{78.7} & \textbf{66.0} & \textbf{64.8} \\
\hline 
\textsc{Teacher} & 81.1 & 83.3 & 66.7 & 65.2 \\
\hline
\end{tabular}
\caption{\label{table:gold-disitillation-results} Gold distillation results ($F_1$ scores).
Best distillation is achieved with $\lambda=1.0$, which represents the student trained only on soft labels from the teacher.
}
\end{table}

To find out if synthetic data can help bridge this gap, we include synthetic training examples in the following experiments.
Since the best distillation setup completely ignores the labels in the training data and relies only on the teacher's soft predictions, we first examine if roundtrip consistency is advantageous for distillation.
We randomly sample two subsets of \textsc{4m} synthetic training examples, one from all generated examples and the other from roundtrip-consistent examples. Let us call the two subsets \textsc{Raw 4m} and \textsc{RC 4m}, respectively.
We train two student models, each for one epoch, using distillation with the two datasets.
As Table~\ref{table:synth-disitillation-rc-results} shows, distillation with \textsc{Raw 4m} demonstrates better results, again likely due to greater sample diversity.
\textsc{Raw 4m} also outperforms gold-only distillation in all four test conditions.

These results suggest that the use of synthetic training examples should be tailored for different applications differently, e.g., RC for pre-training as shown by  \citet{alberti-etal-2019-synthetic} and \textsc{Raw} for distillation as shown in Table~\ref{table:synth-disitillation-rc-results}.
In all our remaining distillation experiments, we use only raw examples generated by our example generator.

To investigate if synthetic distillation can train even better students, we randomly sample subsets of \textsc{1m}, \textsc{2m}, \textsc{4m} and \textsc{6m} raw synthetic training examples.
With each subset, we evaluate two different  distillation strategies: (a) distilling with only synthetic examples (1 epoch), and (b) distilling first with synthetic (1 epoch) and then with gold examples (4 epochs for both SQuAD2.0 and NewsQA, tuned on Dev).
Results for all these student models are shown in Table~\ref{table:synth-disitillation-results}.
Our first observation is that performance consistently improves for both strategies with the amount of synthetic examples used, which supports the hypothesis that synthetic examples in large numbers provide better distillation due to increased input diversity.


Let $\textsc{Dst}^*_S$ and $\textsc{Dst}^*_{S \rightarrow G}$ denote the best-performing student models trained with synthetic-only distillation and synthetic-then-gold distillation, respectively.
In Table~\ref{table:synth-disitillation-results}, we also see that $\textsc{Dst}^*_{S}$ already matches the teacher's performance in two out of the four test conditions (SQuAD2.0 Dev and NewsQA Test).
Fine-tuning with gold distillation further improves performance, surprisingly to the point that $\textsc{Dst}^*_{S \rightarrow G}$ actually outperforms the teacher in three out of the four test conditions (all except NewsQA Dev), by an average margin of about 0.4 point.
As synthetic distillation is computationally expensive, we use a maximum of \textsc{6m} synthetic examples in our experiments; even better results may be achieved with more examples.

\begin{table}[t]
\fontsize{8.6}{10}\selectfont
\centering
\begin{tabular}{lcccc}
\multicolumn{1}{c}{} & \multicolumn{2}{c}{\textbf{SQuAD2.0}} & \multicolumn{2}{c}{\textbf{NewsQA}} \\
\cline{2-5}
& Dev & Test & Dev & Test\\
 \hline
\textsc{Syn RC 4m} & 80.2 & 80.8 & 65.8 & 64.6 \\
\textsc{Syn Raw 4m} & \textbf{80.8} & \textbf{82.1} & \textbf{66.1} & \textbf{65.0} \\
\hline
\textsc{Gld} & 77.4 & 78.7 & 66.0 & 64.8 \\
\hline
\end{tabular}
\caption{\label{table:synth-disitillation-rc-results} Distillation performance ($F_1$ scores) of raw and roundtrip-consistent (RC) synthetic examples.
\textsc{Raw} outperforms both RC and gold examples (\textsc{Gld}).
}
\end{table}

\begin{table}[t]
\fontsize{8.6}{10}\selectfont
\centering
\begin{tabular}{lcccc}
\multicolumn{1}{c}{} & \multicolumn{2}{c}{\textbf{SQuAD2.0}} & \multicolumn{2}{c}{\textbf{NewsQA}} \\
\cline{2-5}
& Dev & Test & Dev & Test\\
 \hline
\textsc{Syn 1m} & 79.0 & 79.9 & 65.0 & 64.0 \\
\textsc{Syn 2m} & 80.0 & 80.9 & 65.7 & 65.0 \\
\textsc{Syn 4m} & 80.8 & 82.1 & 66.1 & 65.0 \\
\textsc{Syn 6m} ($\textsc{Dst}^*_S$) & 81.1 & 82.2 & 66.1 & 65.3 \\

\hline
\textsc{Syn 1m} $\rightarrow$ \textsc{Gld} & 80.1 & 82.2 & 66.3 & 65.1\\
\textsc{Syn 2m} $\rightarrow$ \textsc{Gld} & 80.5 & 82.8 & 66.4 & 65.4\\
\textsc{Syn 4m} $\rightarrow$ \textsc{Gld} & 81.0 & \textbf{83.5} & 66.4 & 65.5\\
\textsc{Syn 6m} $\rightarrow$ \textsc{Gld} ($\textsc{Dst}^*_{S \rightarrow G}$) & \textbf{81.4} & \textbf{83.5} & \textbf{66.6} & \textbf{65.8}\\
\hline 
\textsc{Teacher} & 81.1 & 83.3 & 66.7 & 65.2 \\
\hline
\textsc{BERT-Base Gld MLE} & 74.7 & 75.8 & 62.2 & 60.6 \\
\textsc{Overall} $\Delta$ & +6.7 & +7.7 & +4.4 & +5.2 \\
\hline
\end{tabular}
\caption{\label{table:synth-disitillation-results} Final synthetic distillation results ($F_1$ scores). $\textsc{Dst}^*_S$:~top student with synthetic-only distillation; $\textsc{Dst}^*_{S \rightarrow G}$:~top student with synthetic followed by gold distillation.
$\textsc{Dst}^*_{S \rightarrow G}$ models (\textsc{BERT-Base} students, \textsc{110m} parameters) often outperform their teachers (\textsc{BERT-Large}, \textsc{340m} parameters).
}
\end{table}

\subsection{Summary of Results}
To summarize our main results, we show that highest error synthetic pre-training---where an MRC model (\textsc{BERT-Large} in our experiments) is pre-trained only on the hardest subset of available synthetic examples---outperforms standard pre-training with all synthetic examples on both SQuAD2.0 and NewsQA.
We further show that with a large enough sample of synthetic examples, a much smaller \textsc{BERT-Base} student model can be distilled that not only matches but actually outperforms this strong \textsc{BERT-Large} teacher in most test conditions.
Finally, as shown in Table~\ref{table:synth-disitillation-results} (last two rows), our best student models demonstrate performance increases ranging from 4.4 to 7.7 absolute points over equally sized \textsc{BERT-Base} models trained only on gold examples.

\section{Conclusion}
Light and fast yet high-performance models are a holy grail in practical NLP.
In this work, we take a big step towards achieving this goal for machine reading comprehension (MRC).
Utilizing recent advances in automatic generation of MRC examples, we propose novel applications of synthetic examples that yield large performance improvements over existing techniques.
Crucially, we are able to build smaller models that often outperform larger (over 3$\times$ as many parameters), more powerful models.
Future work will test the limits of the proposed methods, for example with more synthetic examples and ensembles of large models as teachers.
Another important direction is the generalization of these ideas to other tasks for which training examples can be automatically generated.


\bibliography{anthology,acl2020}
\bibliographystyle{acl_natbib}

\appendix

\end{document}